\documentclass[10pt, a4paper]{article}
\usepackage{lrec2022} 
\usepackage{multibib}
\newcites{languageresource}{Language Resources}
\usepackage{graphicx}
\usepackage{tabularx}
\usepackage{soul}
\usepackage{multirow}

\usepackage{multicol}
\usepackage{caption}

\usepackage{titlesec}
\titleformat{\section}{\normalfont\large\bfseries\center}{\thesection.}{1em}{}
\titleformat{\subsection}{\normalfont\SmallTitleFont\bfseries\raggedright}{\thesubsection.}{1em}{}
\titleformat{\subsubsection}{\normalfont\normalsize\bfseries\raggedright}{\thesubsubsection.}{1em}{}
\renewcommand\thesection{\arabic{section}}
\renewcommand\thesubsection{\thesection.\arabic{subsection}}
\renewcommand\thesubsubsection{\thesubsection.\arabic{subsubsection}}

\newcommand{\pd}[1]{$_{\textrm{\scriptsize #1}}$}

\usepackage{epstopdf}
\usepackage[utf8]{inputenc}

\usepackage{hyperref}
\usepackage{xstring}

\usepackage{color}
\usepackage{todonotes}

\title{KIND: an Italian Multi-Domain Dataset for Named-Entity Recognition}

\name{Teresa Paccosi$^{1,2}$, Alessio Palmero Aprosio$^1$}

\address{$^1$ Fondazione Bruno Kessler -- Via Sommarive 18, Trento, Italy \\
         $^2$ Università di Trento -- Corso Bettini 84, Rovereto, Italy \\
         \{tpaccosi, aprosio\}@fbk.eu\\}

\abstract{
In this paper we present KIND, an Italian dataset for Named-entity recognition.
It contains more than one million tokens with annotation covering three classes: person, location, and organization.
The dataset (around 600K tokens) mostly contains manual gold annotations in three different domains (news, literature, and political discourses) and a semi-automatically annotated part.
The multi-domain feature is the main strength of the present work, offering a resource which covers different styles and language uses, as well as the largest Italian NER dataset with manual gold annotations. It represents an important resource for the training of NER systems in Italian. 
Texts and annotations are freely downloadable from the Github repository.
 \\ \newline  }

\begin{document}

\maketitleabstract

\begin{multicols}{2}
\section{Introduction}

Named-entity recognition (NER) is the Natural Language Processing (NLP) task consisting in identifying and classifying mentions of entities in texts.
These mentions belong to a set of predefined categories, among which people, locations, and organizations are the most common.

Like the majority of NLP tasks, especially the ones relying on machine learning algorithms, manually annotated data plays a crucial role, since they are used to train and evaluate the automatic extraction systems that perform the task.
Annotated data are time and money consuming, since they usually need to be created by experts of the domain of the annotation that is going to be done.
While there is plenty of datasets for NER in English, little has been done for other languages, especially for Italian \cite{ehrmann-etal-2016-named}.\footnote{\url{https://damien.nouvels.net/resourcesen/corpora.html}}

The most common general-purpose dataset having named entity annotations is I-CAB \cite{magnini-etal-2006-cab}, created in 2006 and consisting of 525 news stories taken from the local newspaper ``L'Adige'', for a total of around 180,000 words.
Entities in I-CAB are divided into four categories (person, organization, location and geo-political entities), and the dataset is available by signing an agreement and only for research purposes.

The MEANTIME Corpus \citelanguageresource{minard-etal-2016-meantime}, developed within the NewsReader project\footnote{\url{http://www.newsreader-project.eu/}} \cite{VOSSEN201660}, consists of a total of 480 news articles on four topics, taken from Wikinews.
All the topics pertain to the financial domain, accordingly affecting the annotation and selection of the documents.  
The corpus has been annotated manually at multiple levels, including entities, events, temporal information, semantic roles, and intra-document and cross-document event and entity co-reference.

The DBpedia abstract corpus\footnote{\url{http://downloads.dbpedia.org/2015-04/ext/nlp/abstracts/}} \citelanguageresource{Spasojevic:2017:DDA:3041021.3053367} contains a conversion of Wikipedia abstracts in seven languages (including Italian), with the annotations of linked entities, manually disambiguated to Wikipedia/DBpedia resources by native speakers.
Similarly, the DAWT dataset includes 13.6 million articles extracted from Wikipedia with labeled text mentions mapping to entities as well as the type of the entity.

Finally, WikiNER \citelanguageresource{Nothman2013LearningMN} is a free and multilingual dataset for NER by exploiting the text and structure of Wikipedia.
Annotations from the last three datasets rely on the Wikipedia links, whose coverage was improved by using different strategies.
For this reason, data included can be considered as silver-standard.


In this paper, we present an Italian dataset called KIND (Kessler Italian Named-entities Dataset) annotated with named entities belonging to three classes (person, location, organization), containing more than one million tokens, among with 600K are manually annotated.
The main strength of the present corpus is to be multi-domain, since it contains news articles, literature, and political speeches.
KIND is available under the Attribution-NonCommercial 4.0 International (CC BY-NC 4.0) license, and it is freely downloadable from Github.

In Section~\ref{sec:corpus} we describe how the texts are collected, while in Section~\ref{sec:annotation} we outline the annotation process.
In Section~\ref{sec:evaluation} we evaluate the dataset and discuss about some outcomes.
Finally, Section~\ref{sec:release} contains information on how to download and use KIND.


\section{Description of the Corpus}
\label{sec:corpus}

\begin{table*}
\begin{center}
{\setlength{\tabcolsep}{0.4em}\small
\begin{tabular}{|l|r|r|r|r|r|r|r|r|r|r|}

      \hline
      \multirow{2}{*}{Dataset} & \multirow{2}{*}{Documents} & \multirow{2}{*}{Tokens} & \multicolumn{4}{c|}{Train} & \multicolumn{4}{c|}{Test} \\
      \cline{4-11}
      & & & Total & PER & ORG & LOC & Total & PER & ORG & LOC \\
      \hline
      \hline
      Wikinews & 1,000 & 308,622 & 247,528 & 8,928 & 7,593 & 6,862 & 61,094 & 1,802 & 1,823 & 1,711\\
      \hline
      Fiction & 86 & 192,448 & 170,942 & 3,439 & 182 & 733 & 21,506 & 636 & 284 & 463 \\
      \hline
      Aldo Moro & 250 & 392,604 & 309,798 & 1,459 & 4,842 & 2,024 & 82,806 & 282 & 934 & 807\\
      \hline
      Alcide De Gasperi & 158 & 150,632 & 117,997 & 1,129 & 2,396 & 1,046 & 32,635 & 253 & 533 & 274\\
      \hline
      \hline
      Total & 1,494 & 1,044,306 & 846,265 & 14,955 & 15,013 & 10,665 & 198,041 & 2,973 & 3,574 & 3,255 \\
      \hline

\end{tabular}
\caption{Overview of the dataset}
\label{tab:overview}}
\end{center}
\end{table*}

For the construction of the dataset, we decide to use texts available in the public domain, under a license that allows both research and commercial use.
In particular we release four chapters with texts taken from: (i) Wikinews (WN) as a source of news texts from the last 20 years; (ii) some Italian fiction books (FIC) 
publicly available; (iii) writings and speeches from Italian politicians Aldo Moro (AM) and (iv) Alcide De Gasperi (ADG).

Apart from Aldo Moro's writings (see Section~\ref{sec:morosilver}), all the annotations are entirely manually tagged by expert linguists .

Table~\ref{tab:overview} shows an overview of the content of the different datasets included in KIND.

\subsection{Wikinews}
\label{sec:wikinews}

Wikinews is a multi-language free-content project of collaborative journalism.
The Italian chapter contains more than 11,000 news articles,\footnote{\url{https://it.wikinews.org/wiki/Speciale:Statistiche}} released under the Creative Commons Attribution 2.5 License.\footnote{\url{https://creativecommons.org/licenses/by/2.5/}}

In building KIND, we randomly choose 1,000 articles evenly distributed in the last 20 years, for a total of 308,622 tokens.

\subsection{Literature}
\label{sec:fiction}

Regarding fiction literature, we annotate 86 book chapters from 10 books written by Italian authors, whose works are publicly available, for a total of 192,448 tokens.
The selected books are mostly novels, but there are also epistles and biographies. 
The plain texts are taken from the Liber Liber website.\footnote{\url{https://www.liberliber.it/}}

In particular, we select: \textit{Il giorno delle Mésules} (Ettore Castiglioni, 1993, 12,853 tokens), \textit{L'amante di Cesare} (Augusto De Angelis, 1936, 13,464 tokens), \textit{Canne al vento} (Grazia Deledda, 1913, 13,945 tokens), \textit{1861-1911 - Cinquant’anni di vita nazionale ricordati ai fanciulli} (Guido Fabiani, 1911, 10,801 tokens), \textit{Lettere dal carcere} (Antonio Gramsci, 1947, 10,655), \textit{Anarchismo e democrazia} (Errico Malatesta, 1974, 11,557 tokens), \textit{L'amore negato} (Maria Messina, 1928, 31,115 tokens), \textit{La luna e i falò} (Cesare Pavese, 1950, 10,705 tokens), \textit{La coscienza di Zeno} (Italo Svevo, 1923, 56,364 tokens), \textit{Le cose più grandi di lui} (Luciano Zuccoli, 1922, 20,989 tokens).

In selecting works which are in the public domain, we favored texts as recent as possible, so that the model trained on this data would be efficiently applied to novels written in the last years, since the language used in these novels is more likely to be similar to the language used in the novels of our days.

\subsection{Aldo Moro's Works}
\label{sec:moro}

Writings belonging to Aldo Moro have recently been collected by the University of Bologna and published on a platform called ``Edizione Nazionale delle Opere di Aldo Moro'' \citelanguageresource{aldomorodigitale}.\footnote{\url{https://aldomorodigitale.unibo.it/}}
The project is still ongoing and, by now, it contains 806 documents for a total of about one million tokens.

In the first release of KIND, we include 392,604 tokens from the Aldo Moro's works dataset, with silver annotations (see Section~\ref{sec:morosilver}).

\subsection{Alcide De Gasperi's Writings}
\label{sec:degasperi}

Finally, we annotate 158 document (150,632 tokens) from the corpus described in \citelanguageresource{Tonelli2019PrendoLP}, spanning 50 years of European history.
The corpus is composed of a comprehensive collection of Alcide De Gasperi’s public documents, 2,762 in total, written or transcribed between 1901 and 1954, and it is available for consultation on the Alcide Digitale website.\footnote{\url{https://alcidedigitale.fbk.eu/}}

\section{Annotation Process}
\label{sec:annotation}

\subsection{Data preprocessing}
\label{sec:preprocessing}

To annotate the documents, we start from plain texts.
For Wikinews, all the texts are extracted from the dumps server\footnote{\url{https://dumps.wikimedia.org/}} and cleaned from markup (e.g. text formatting) using the Bliki engine.\footnote{\url{https://github.com/axkr/info.bliki.wikipedia_parser}}
Documents from other sources (Aldo Moro's and Alcide De Gasperi's writings, along with fiction books) are already in plain text format, therefore they do not need any particular conversion.

Finally, tokens and sentences are extracted using the Tint NLP Suite \cite{alessio2018tint}.


\subsubsection{Aldo Moro's writings silver data}
\label{sec:morosilver}

As part of the activities of the project ``Edizione Nazionale delle Opere di Aldo Moro'', entities in documents from Aldo Moro's dataset have already been identified by a group of expert annotators, mainly historians and archivists.
Therefore, most of the named entities can be extracted and tagged as person, location or organization without any additional manual effort.
However, the guidelines for the annotation defined within the project are different from ours.
For example, common nouns such as ``ministro'' are tagged and linked to the referred person, even when the person's name is not included in the document (but can be inferred by the context).
For this reason, we perform a semi-automatic check of the whole corpus, following these steps:

\begin{itemize}
    \item the complete set of unique entities have been extracted from the corpus and manually checked by an expert;
    \item the entities confirmed as entities in the previous step are then searched and automatically tagged (case sensitive) in the corpus;
    \item finally, the corpus is processed with the NER module of Tint and the additional entities found in this phase are manually checked by an expert (and eventually added to the annotation).
\end{itemize}

By the application of the above-described steps, we enhance the compliance of the chapter with our guidelines.
Nevertheless, we prefer to call the annotations of this part of the dataset ``silver'' (and not ``gold''), because they are a mix of manually and automatically tagged entities.
See Section~\ref{sec:discussion} for more information.

\subsection{Annotation Tool}
To carry out the annotation of entities in KIND we choose INCEpTION \footnote{\url{https://inception-project.github.io/}} \cite{tubiblio106270} , a web-based text-annotation environment which turns out to be the most suitable for the task, since it presents several advantages. 
In fact, the tool is endowed with an intuitive environment in which the labels are fully customisable. 
It also includes automatic label propagation functionality, which speeds up the process of annotation conspicuously. 
According to the tagging scheme we choose for the annotation (IOB), INCEpTION results as the most suitable choice also because it gives us the possibility to link the annotated spans among them, so that in a case such as ``[Paolo] [Rossi] è andato a Roma'' (``[Paolo] [Rossi] went to Rome'') I can link the name [Paolo] to the surname [Rossi], which successively will be labelled respectively as B-PER, I-PER. 
 


\subsection{Annotation Tagging Scheme}

As said above, we used the Inside-Outside-Begining (IOB) tagging scheme, which subdivides the elements of every entity as begin-of-entity (B-ent) or continuation-of-entity (I-ent). According to this format, the entities are marked such as the first element of the ``compound'' entity is B-ent and the following ones are I-ents, as we can see in the sentence: ``La chiameranno [Fondazione]\pd{B-ORG} [Bruno]\pd{I-ORG} [Kessler]\pd{I-ORG}'' (``They will call it, [Bruno] [Kessler] [Foundation]''), where B stands for Beginning and I for Inside. 
This holds true for all the named entities, such as PER (``[Sophia]\pd{B-PER} [Loren]\pd{I-PER}, famosa attrice italiana'', ``[Sophia] [Loren], famous italian actress'') or LOC ([``Via]\pd{B-LOC} [Nazionale]\pd{I-LOC} [12]\pd{I-LOC}''). 
When the entity is composed of only one element, the annotation scheme treats it as it would be a first element of a compound one, with B-ent (``La rassegna è stata promossa dal [CNR]\pd{B-ORG}'', ``the exhibition was promoted by [CNR]''). 
We choose to not consider nested entities as a different case (such as ``Fondazione Bruno Kessler" which it is an ORG which contains a PER entity) but to annotate only the element considered in the sentence in which it is contain. 
For instance, in the sentence ``Lavora per la [Fondazione Bruno Kessler]"(``He/she works for the [Fondazione Bruno Kessler]"), ``Fondazione Bruno Kessler" would be annotated only as ORG entity.

\subsection{Person Entities}

We consider as PERSON ENTITIES (PER) those entities which refer to an individual or an animal by his/her proper name, such as in the following sentences (extracted by the present corpus) where PER is contained in square brackets: ``[Laura] vive a Roma'' (``[Laura] lives in Rome''); ``Partecipa la Torre con [Tremendo] su [Guess]'' (``The Tower participates with [Tremendo] on [Guess]'', where ``Guess'' is the horse and ``Tremendo'' the jockey). 
We annotate as PER also fictitious characters, as long as they possess a proper name (e.g. Mickey Mouse), as well as proper names which refer to a group of people belonging to the same family (e.g. the Jackson). In the case there is an apposition, a title or a function preceding the proper person name, since we consider only proper names, we annotate only the proper noun and not the common noun associated, as in the case of ``papa [Giovanni] [Paolo] [II] ha viaggiato molto'' (``pope [John] [Paul] [II] has travelled a lot''). 

\subsection{Organization Entities}

We consider as ORGANIZATION ENTITIES (ORG) every formally established association. 
These associations can be of different types such as governmental (``Il [governo] [italiano] si è espresso a favore'', ``the [italian] [government] had spoken in favour''), commercial (``I ricavi di [Zoom] sono incrementati'', ``the revenues of [Zoom] has increased''), educational (``L’[Università] [di] [Pisa] avrà un nuovo rettore'', ``the [University] [of] [Pisa] will have a new rector''), related to media (``Si tratta di Domenico Quirico de [La] [Stampa]'', ``It is Domenico Quirico, from [La] [Stampa]''), religious (``Le posizioni della [Chiesa]  non hanno alcuna collocazione politica'', ``the views of the [Church] have no political placement''), related to sports ([Juve] - [Roma] 1 – 1), medical-scientific (``Il giovane venne soccorso dall’elicottero del [118]'', ``the boy was rescued by the helicopter of [118]''), non-governmental (such as political parties, professional regulatory or no-profit organizations), related to entertainment (``I [R.E.M.] presenteranno il loro nuovo album'', ``[R.E.M.] will present their new album''), and also brands names in general (``Creò anche la linea [Chicco] per [Artsana]'', ``He created also the [Chicco] line for [Artsana]''. 
In general we can say that every time the entity is the agent of an action and it cannot be defined as a PER, the label to choose for the annotation is ORG. We will address the issue more in depth in ~\ref{sec:discussion}.

\subsection{Location Entities}

We consider as LOCATION ENTITIES (LOC) those entities referring to places defined on a geographical basis or, more in general, entities which possess a physical location and a proper name. 
Therefore, we annotate as LOC nations, continents, cities but also facilities or shops, bar and restaurants if they have a proper name (e.g. ``[Torre] [Eiffel]''; ``[Bar] [Il] [Giobertino]''; ``[Stazione] [Termini]''). According to this rule, in the sentence ``Sono nato all'[ospedale] [Torregalli]'' (``I was born at [Torregalli] [hospital]'') has to be annotated, since ``Torregalli'' is the name of the hospital, but in the sentence ``l'ospedale di [Firenze] si trova in fondo a questa strada'' (``the hospital of [Florence] is a the end of this street'') only ``Firenze'' has to be annotated, as Florence is the name of the city but the hospital does not present a proper name. 
We also annotate as LOC entities those places contained in a prize name or in the name of a race/competition when the award ceremony or race/competition takes place in the location described, such as in ``Premio [Roma]'', ``Coppa [Italia]'' or ``Rally [Dakar]''.

\subsection{Annotation Guidelines}

In the previous sections we described the annotation process of all the classes in detail. Beyond the above-cited examples, the annotators found some uncertain cases. 
On GitHub we provide the annotation guidelines for all the classes, with dedicated paragraphs for the ambiguous cases that the annotators found, such as metonymy or the inclusion of titles and functions in the annotation of entities. 

\section{Evaluation}
\label{sec:evaluation}

To evaluate the accuracy of the annotation, we use different metrics.
First of all, through inter-annotator agreement we check the effectiveness of the guidelines.
Then, we train two NER models using different algorithms: Conditional Random Fields \cite{laffertyCrf} and BERT \cite{devlin2019bert}.
The former is more common and widely used in production environments since it can be trained on traditional CPUs, while the latter represents the state-of-the-art, but it usually requires GPUs to reduce classification time.

\subsection{Inter-Annotator Agreement}

The most common way to measure inter-annotator agreement is the Cohen's kappa statistic $\kappa$ \cite{cohenskappa}, which takes into account the amount of agreement that could be expected to occur through chance.

To compute this value, two different  expert linguists are asked to annotate the same set of 15 documents randomly extracted from Wikinews.
We obtain $\kappa=0.952$, meaning an excellent agreement.

\begin{table*}
\begin{center}
{\setlength{\tabcolsep}{0.35em}\small
\begin{tabular}{|l|l|l||r|r|r||r|r|r||r|r|r||r|r|r||r|r|r|}
      \hline
      \multirow{2}{*}{Algo} & \multicolumn{2}{c||}{Dataset} &
        \multicolumn{3}{c||}{PER} & \multicolumn{3}{c||}{LOC} &
        \multicolumn{3}{c||}{ORG} &
        \multicolumn{3}{c||}{Micro} & \multicolumn{3}{c|}{Macro} \\
      \cline{2-18}
      & Train & Test & P & R & $F_1$ & P & R & $F_1$ & P & R & $F_1$
        & P & R & $F_1$ & P & R & $F_1$ \\
      \hline
      \hline
CRF & WN & WN & 0.91 & 0.92 & 0.92 & 0.85 & 0.82 & 0.83 & 0.79 & 0.71 & 0.75 & 0.85 & 0.82 & 0.83 & 0.85 & 0.82 & 0.83 \\
      \hline
CRF & AM & AM & 0.97 & 0.91 & 0.94 & 0.96 & 0.97 & 0.96 & 0.93 & 0.94 & 0.94 & 0.95 & 0.95 & 0.95 & 0.95 & 0.94 & 0.95 \\
CRF & ADG & AM & 0.95 & 0.79 & 0.86 & 0.94 & 0.62 & 0.74 & 0.61 & 0.77 & 0.68 & 0.74 & 0.71 & 0.73 & 0.83 & 0.72 & 0.76 \\
CRF & ADG & ADG & 0.92 & 0.88 & 0.90 & 0.87 & 0.69 & 0.77 & 0.80 & 0.67 & 0.73 & 0.85 & 0.72 & 0.78 & 0.86 & 0.75 & 0.80 \\
CRF & AM & ADG & 0.91 & 0.80 & 0.85 & 0.72 & 0.72 & 0.72 & 0.90 & 0.41 & 0.57 & 0.84 & 0.58 & 0.69 & 0.84 & 0.64 & 0.71 \\
      \hline
CRF & FIC & FIC & 0.81 & 0.77 & 0.79 & 0.61 & 0.76 & 0.68 & 0.74 & 0.25 & 0.37 & 0.72 & 0.66 & 0.69 & 0.72 & 0.59 & 0.61 \\
CRF & WN & FIC & 0.89 & 0.72 & 0.80 & 0.71 & 0.80 & 0.75 & 0.63 & 0.68 & 0.65 & 0.76 & 0.74 & 0.75 & 0.74 & 0.73 & 0.73 \\
CRF & WN+FIC & FIC & 0.90 & 0.78 & 0.84 & 0.73 & 0.81 & 0.77 & 0.70 & 0.66 & 0.68 & 0.79 & 0.76 & 0.78 & 0.78 & 0.75 & 0.76 \\
    \hline
      \hline
BERT & WN & WN & 0.96 & 0.96 & 0.96 & 0.88 & 0.90 & 0.89 & 0.83 & 0.82 & 0.82 & 0.89 & 0.89 & 0.89 & 0.89 & 0.89 & 0.89 \\
      \hline
BERT & AM & AM & 0.97 & 0.97 & 0.97 & 0.93 & 0.97 & 0.95 & 0.86 & 0.94 & 0.90 & 0.90 & 0.96 & 0.93 & 0.92 & 0.96 & 0.94 \\
BERT & ADG & AM & 0.93 & 0.92 & 0.92 & 0.90 & 0.54 & 0.68 & 0.53 & 0.85 & 0.65 & 0.66 & 0.74 & 0.69 & 0.79 & 0.77 & 0.75 \\
BERT & ADG & ADG & 0.96 & 0.88 & 0.91 & 0.86 & 0.83 & 0.85 & 0.75 & 0.77 & 0.76 & 0.82 & 0.81 & 0.82 & 0.86 & 0.83 & 0.84 \\
BERT & AM & ADG & 0.92 & 0.86 & 0.89 & 0.75 & 0.80 & 0.78 & 0.87 & 0.52 & 0.65 & 0.84 & 0.68 & 0.75 & 0.85 & 0.73 & 0.77 \\
      \hline
BERT & FIC & FIC & 0.94 & 0.93 & 0.94 & 0.76 & 0.85 & 0.80 & 0.77 & 0.41 & 0.54 & 0.84 & 0.80 & 0.82 & 0.82 & 0.73 & 0.76 \\
BERT & WN & FIC & 0.94 & 0.94 & 0.94 & 0.81 & 0.89 & 0.85 & 0.69 & 0.81 & 0.75 & 0.84 & 0.90 & 0.87 & 0.81 & 0.88 & 0.84 \\
BERT & WN+FIC & FIC & 0.94 & 0.94 & 0.94 & 0.81 & 0.88 & 0.84 & 0.75 & 0.85 & 0.80 & 0.85 & 0.90 & 0.88 & 0.83 & 0.89 & 0.86 \\
      \hline
\end{tabular}
}
\caption{Results of training on the KIND dataset}
\label{tab:results}
\end{center}
\end{table*}

\subsection{Experiments}
\label{sec:experiments}


Before running the experiments to train the models for the task of NER, we split the different datasets into train and test.
Typically, documents are shuffled and a subset of them is extracted to reach around 20\% of the total.
Table~\ref{tab:overview} shows how the partitioning is performed in the chapters of the dataset.
The main purpose of the fiction chapter is the possibility to train a model for NER that can be applied efficiently to literary texts in general. 
For this reason, we choose to not extract the test sentences randomly in the chapter, but to use as test set the works of two authors (Guido Fabiani and Cesare Pavese), who clearly are not included in the training set, in order to avoid the possibility to have a model trained on a particular writing style. 
This would allow to have a more realistic result in terms of performance.


To train the Conditional Random Fields (CRF) model, we run the implementation included in Stanford CoreNLP \cite{manning-EtAl:2014:P14-5}, already used in its NER module \cite{finkel-etal-2005-incorporating}.

We tries some sets of features choosing among the ones available in the software.
We obtain the best results with word shapes, n-grams with length 6, previous, current, and next token/lemma/class.

To enhance the classification, Stanford NER also accepts gazetteers of names labelled with the corresponding tag.
We collect a list of persons, organizations and locations from the Italian Wikipedia using some classes in DBpedia \cite{10.1007/978-3-540-76298-0_52}: \texttt{Person}, \texttt{Organization}, and \texttt{Place}, respectively.
Table~\ref{tab:gazette} shows statistics about the gazettes.

For BERT, we use an adaptation of the Bert Model with a token classification head on top, available in the transformers Python package\footnote{\url{https://github.com/huggingface/transformers}} developed by Hugging Face.\footnote{\url{https://huggingface.co/huggingface}}
The model is trained (3 epochs are enough) starting from the \texttt{bert-base-italian-cased} model.\footnote{\url{https://huggingface.co/dbmdz/bert-base-italian-cased}}

Table~\ref{tab:results} shows classification results comparing different configurations and algorithms.

\begin{center}
\begin{tabular}{|l|l|r|}
\hline
\bf Source & \bf Tag & \bf Labels \\
\hline
Wikipedia & LOC & 377,611 \\
Wikipedia & PER & 608,547 \\
Wikipedia & ORG & 84,887 \\
\hline
\end{tabular}
\captionof{table}{ Items added to the CRF NER training taken from gazettes. }
\label{tab:gazette}
\end{center}

\subsection{Discussion}
\label{sec:discussion}

We have seen above that LOC and ORG indicate different entities but there are some cases in which these two labels can overlap. 
This is the case of metonymic location names, a special case which occurs when the proper name of a LOC entity is used to refer to an organization.
The most common case is when the name of a location is used to refer to a sport team (``La [Russia] ha conquistato la medaglia d’oro'', ``[Russia] have won the gold medal'') but, more in general, we can say that the cases in which the LOC entity is treated as an ORG entity include all the situations in which the LOC entity is the agent of an action, such as in the sentence ``La [Germania] si ritira dalla trattativa'' (``Germany withdrew from negotiations''), where Germany is the subject of the action of withdrawal, and it is then indicated as ORG, or ``La [Lombardia] si dice contraria al provvedimento'' (``[Lombardy] declares to be contrary to the measure''), where Lombardy is the subject as governamental body.

\subsubsection{LOC annotations in AM chapter}

As said in Section~\ref{sec:morosilver}, the chapter of KIND containing documents from Aldo Moro's work were already annotated with some named entities.
In particular, countries and regions were almost all tagged correctly and even linked to the corresponding Wikidata page.
Unfortunately, the policies used by the annotators is slightly different from the one adopted in KIND, therefore all the locations are tagged as LOC, even when they represent an organization (and therefore should be tagged as ORG in KIND).
For this reason, we performed some additional experiments using ADG and AM as train and test, respectively.
As one can see, while both models are very accurate on PER, precision and recall on LOC and ORG drop at around 0.5 in some cases.
In particular, when training on ADG and testing on AM, recall on LOC and precision on ORG are very low, while switching the datasets (training on AM and testing on ADG) will result in a low recall on ORG.
This is a result of the different policy in annotating the two datasets.

\subsubsection{Some comments about the fiction dataset}

Regarding experiments on fictional texts (FIC) test set (see Section~\ref{sec:experiments}), we try three different configurations: FIC train set, WN train set, both train sets.

In the first configuration (FIC), one can see that classification of PER and LOC is quite good, while recall on ORG is really low: this is due to the scarcity of organization mentions in fictional texts.
In the second experiment (where WN has been used as training set), the recall on ORG is higher, but the precision drops almost 10 points.
This is probably motivated by the different contexts in which organizations are mentioned in literature, w.r.t. news.
One can see that on PER and LOC tags the accuracy is higher, meaning that usually there are no relevant differences in uses in literature and news for these two categories of entities.
We finally merge the training sets of both datasets, resulting in a classification where accuracy is comparable to the one where WN is used for testing.
As a hint, we suggest to use both datasets (expecially for a correct ORG identification) when one needs to tag fictional texts.


\section{Release}
\label{sec:release}

The KIND dataset is released in open access on Github.\footnote{\url{https://github.com/dhfbk/KIND}}
Some supporting tools (such as Wikinews extraction and format conversion tools) are part of the \texttt{tint-resources} package included in Tint release.\footnote{\url{https://github.com/dhfbk/tint}}

The texts used for KIND are available with different licenses (see the single dataset description for more information), but all of them are available to use for non-commercial purposes.

The named entities annotations in WN, FIC, and AM are released under the Attribution-NonCommercial 4.0 International (CC BY-NC 4.0) license.\footnote{\url{https://creativecommons.org/licenses/by-nc/4.0/}}
Annotations in ADG are released under the Attribution-NonCommercial-ShareAlike 4.0 International (CC BY-NC-SA 4.0) license.\footnote{\url{https://creativecommons.org/licenses/by-nc-sa/4.0/}}


\section{Conclusion and Future Work}
\label{sec:conclusion}

In this paper we presented KIND, a multi-domain dataset in Italian containing more then one million tokens annotated with named entities (person, location, organization).
Part of the dataset (more than 600K tokens) contains gold annotations, manually performed by  expert linguists.
The remaining part includes around 400K tokens from Aldo Moro's works and is annotated by automatically converting an existing annotation, built using different guidelines and for different purposes.

In future, we plan to increase the number of documents, especially in the chapter that contains literature texts.
We would also like to include all the texts from Aldo Moro's works, once they are available.

\section*{Acknowledgements}

The authors thank Arianna Pergher, a student who helped during the first phase of the annotations process.


\section*{Bibliographical References}\label{reference}

\bibliographystyle{lrec2022-bib}
\bibliography{bibliography}

\begin{thebibliography}{}

\bibitem[\protect\citename{Barzaghi and Paolucci}2021]{aldomorodigitale}
Barzaghi, Sebastian and Paolucci, Francesco.
\newblock (2021).
\newblock {\em {Edizione Nazionale delle Opere di Aldo Moro (Dataset RDF)}}.

\bibitem[\protect\citename{Minard \bgroup et al.\egroup
  }2016]{minard-etal-2016-meantime}
Minard, A.-L., Speranza, M., Urizar, R., Altuna, B., van Erp, M., Schoen, A.,
  and van Son, C.
\newblock (2016).
\newblock {MEANTIME}, the {N}ews{R}eader multilingual event and time corpus.
\newblock In {\em Proceedings of the Tenth International Conference on Language
  Resources and Evaluation ({LREC}'16)}, pages 4417--4422, Portoro{\v{z}},
  Slovenia, May. European Language Resources Association (ELRA).

\bibitem[\protect\citename{Nothman \bgroup et al.\egroup
  }2013]{Nothman2013LearningMN}
Nothman, J., Ringland, N., Radford, W., Murphy, T., and Curran, J.~R.
\newblock (2013).
\newblock Learning multilingual named entity recognition from wikipedia.
\newblock {\em Artif. Intell.}, 194:151--175.

\bibitem[\protect\citename{Spasojevic \bgroup et al.\egroup
  }2017]{Spasojevic:2017:DDA:3041021.3053367}
Spasojevic, N., Bhargava, P., and Hu, G.
\newblock (2017).
\newblock Dawt: Densely annotated wikipedia texts across multiple languages.
\newblock In {\em Proceedings of the 26th International Conference on World
  Wide Web Companion}, WWW '17 Companion, pages 1655--1662, Republic and Canton
  of Geneva, Switzerland. International World Wide Web Conferences Steering
  Committee.

\bibitem[\protect\citename{Tonelli \bgroup et al.\egroup
  }2019]{Tonelli2019PrendoLP}
Tonelli, S., Sprugnoli, R., and Moretti, G.
\newblock (2019).
\newblock Prendo la parola in questo consesso mondiale: A multi-genre 20th
  century corpus in the political domain.
\newblock In {\em CLiC-it}.

\end{thebibliography}


\begin{thebibliography}{}

\bibitem[\protect\citename{Auer \bgroup et al.\egroup
  }2007]{10.1007/978-3-540-76298-0_52}
Auer, S., Bizer, C., Kobilarov, G., Lehmann, J., Cyganiak, R., and Ives, Z.
\newblock (2007).
\newblock Dbpedia: A nucleus for a web of open data.
\newblock In Karl Aberer, et~al., editors, {\em The Semantic Web}, pages
  722--735, Berlin, Heidelberg. Springer Berlin Heidelberg.

\bibitem[\protect\citename{Cohen}1960]{cohenskappa}
Cohen, J.
\newblock (1960).
\newblock A coefficient of agreement for nominal scales.
\newblock {\em Educational and Psychological Measurement}, 20(1):37--46.

\bibitem[\protect\citename{Devlin \bgroup et al.\egroup }2019]{devlin2019bert}
Devlin, J., Chang, M.-W., Lee, K., and Toutanova, K.
\newblock (2019).
\newblock Bert: Pre-training of deep bidirectional transformers for language
  understanding.

\bibitem[\protect\citename{Ehrmann \bgroup et al.\egroup
  }2016]{ehrmann-etal-2016-named}
Ehrmann, M., Nouvel, D., and Rosset, S.
\newblock (2016).
\newblock Named entity resources - overview and outlook.
\newblock In {\em Proceedings of the Tenth International Conference on Language
  Resources and Evaluation ({LREC}'16)}, pages 3349--3356, Portoro{\v{z}},
  Slovenia, May. European Language Resources Association (ELRA).

\bibitem[\protect\citename{Finkel \bgroup et al.\egroup
  }2005]{finkel-etal-2005-incorporating}
Finkel, J.~R., Grenager, T., and Manning, C.
\newblock (2005).
\newblock Incorporating non-local information into information extraction
  systems by {G}ibbs sampling.
\newblock In {\em Proceedings of the 43rd Annual Meeting of the Association for
  Computational Linguistics ({ACL}{'}05)}, pages 363--370, Ann Arbor, Michigan,
  June. Association for Computational Linguistics.

\bibitem[\protect\citename{Klie \bgroup et al.\egroup }2018]{tubiblio106270}
Klie, J.-C., Bugert, M., Boullosa, B., de~Castilho, R.~E., and Gurevych, I.
\newblock (2018).
\newblock {The INCEpTION Platform: Machine-Assisted and Knowledge-Oriented
  Interactive Annotation}.
\newblock In {\em Proceedings of the 27th International Conference on
  Computational Linguistics: System Demonstrations}, pages 5--9. Association
  for Computational Linguistics.

\bibitem[\protect\citename{Lafferty \bgroup et al.\egroup }2001]{laffertyCrf}
Lafferty, J.~D., McCallum, A., and Pereira, F. C.~N.
\newblock (2001).
\newblock {Conditional Random Fields: Probabilistic Models for Segmenting and
  Labeling Sequence Data}.
\newblock In {\em Proceedings of the Eighteenth International Conference on
  Machine Learning}, ICML '01, pages 282--289, San Francisco, CA, USA. Morgan
  Kaufmann Publishers Inc.

\bibitem[\protect\citename{Magnini \bgroup et al.\egroup
  }2006]{magnini-etal-2006-cab}
Magnini, B., Pianta, E., Girardi, C., Negri, M., Romano, L., Speranza, M.,
  Bartalesi~Lenzi, V., and Sprugnoli, R.
\newblock (2006).
\newblock {I}-{CAB}: the {I}talian content annotation bank.
\newblock In {\em Proceedings of the Fifth International Conference on Language
  Resources and Evaluation ({LREC}{'}06)}, Genoa, Italy, May. European Language
  Resources Association (ELRA).

\bibitem[\protect\citename{Manning \bgroup et al.\egroup
  }2014]{manning-EtAl:2014:P14-5}
Manning, C.~D., Surdeanu, M., Bauer, J., Finkel, J., Bethard, S.~J., and
  McClosky, D.
\newblock (2014).
\newblock The {Stanford} {CoreNLP} natural language processing toolkit.
\newblock In {\em Association for Computational Linguistics (ACL) System
  Demonstrations}, pages 55--60.

\bibitem[\protect\citename{Palmero~Aprosio and Moretti}2018]{alessio2018tint}
Palmero~Aprosio, A. and Moretti, G.
\newblock (2018).
\newblock Tint 2.0: an all-inclusive suite for nlp in italian.
\newblock In {\em Proceedings of the Fifth Italian Conference on Computational
  Linguistics CLiC-it}, volume~10, page~12.

\bibitem[\protect\citename{Vossen \bgroup et al.\egroup }2016]{VOSSEN201660}
Vossen, P., Agerri, R., Aldabe, I., Cybulska, A., {van Erp}, M., Fokkens, A.,
  Laparra, E., Minard, A.-L., {Palmero Aprosio}, A., Rigau, G., Rospocher, M.,
  and Segers, R.
\newblock (2016).
\newblock Newsreader: Using knowledge resources in a cross-lingual reading
  machine to generate more knowledge from massive streams of news.
\newblock {\em Knowledge-Based Systems}, 110:60--85.

\end{thebibliography}

\section*{Language Resource References}
\label{lr:ref}
\bibliographystylelanguageresource{lrec2022-bib}
\bibliographylanguageresource{languageresource}

\end{multicols}
\end{document}